\documentclass[conference]{IEEEtran}
\IEEEoverridecommandlockouts

\usepackage{amsmath,amssymb}
\usepackage{graphicx}
\usepackage{booktabs}
\usepackage{multirow}
\usepackage{cite}
\usepackage{balance}
\usepackage{microtype}
\usepackage{algorithm}
\usepackage{algorithmic}
\usepackage{tikz}
\usetikzlibrary{arrows.meta,positioning,fit}
\usepackage{xcolor}

\title{ORBITAL: Optimization of Robust Brain--Computer Inference Through Architecture-Aware Low-Precision Compression}

\author{\IEEEauthorblockN{Saim Rehman and Muhammad Shafique}
\IEEEauthorblockA{eBRAIN Lab, New York University Abu Dhabi, Abu Dhabi, UAE} sr7849@nyu.edu, muhammad.shafique@nyu.edu}

\begin{document}
\maketitle

\begin{abstract}
Compact electroencephalography (EEG) decoders enable embedded brain--computer interfaces (BCIs), but compression can alter adversarial robustness without proportional hardware savings. We present ORBITAL, a robustness-constrained framework combining hardware-realizable structured compression, recovery-aware operating-point selection, multi-attack evaluation, and native edge validation. Across nine subjects and three seeds on BCI Competition IV-2a, S25-TRADES achieves the highest observed mean true-union robustness among compressed EEGNet models (23.06\% vs 21.27\% for standard Full), with 26.27\% fewer parameters and 25.10\% fewer MACs. Matched controls identify TRADES as the main source of robustness recovery, improving S25 union accuracy by 6.74 percentage points over clean recovery (p=0.0039), while per-example analysis reveals boundary redistribution rather than uniform margin expansion. On Jetson Orin Nano, FP16 compression reduces board energy by up to 16.20\% and latency by 8.00\%. Despite 99.3\% prediction agreement, native INT8 Q/DQ incurs about twice the FP16 latency and energy, demonstrating the need for joint robustness and hardware-aware optimization.
\end{abstract}


\section{Introduction}

EEG-based BCIs increasingly rely on compact convolutional neural networks such as EEGNet~\cite{lawhern2018eegnet} and ShallowConvNet~\cite{schirrmeister2017deep}. Their embedded use creates three coupled objectives: (i) preserve task accuracy, (ii) resist small adversarial perturbations, and (iii) reduce real latency/energy. Conventional compression work largely optimizes storage or arithmetic cost~\cite{han2016deep,li2017pruning,molchanov2017pruning,jacob2018quantization}, whereas adversarial training and evaluation are usually studied independently~\cite{madry2018towards,zhang2019trades,croce2020reliable}. In EEG, this separation is particularly problematic because compact decoders operate on low-amplitude, non-image signals for which perturbation geometry and hardware overheads differ from standard vision workloads.

\begin{figure}[t]
    \centering
    \includegraphics[width=\columnwidth]{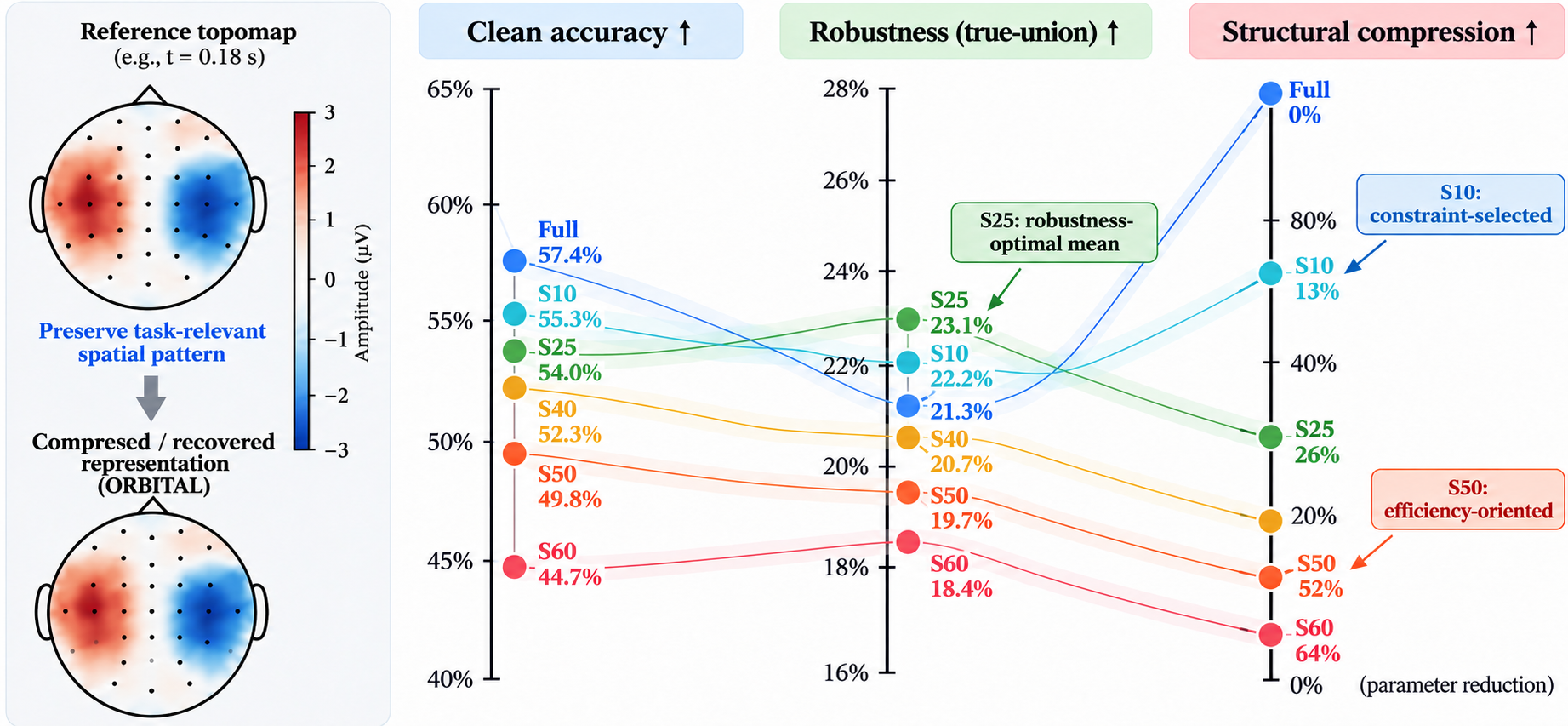}
    \caption{Motivation for robustness-constrained EEG compression. ORBITAL's accuracy--robustness--hardware trade-off under structured EEG compression. S10 satisfies predefined screening constraints, S25 achieves the highest observed mean true-union robustness among compressed candidates, and S50 offers greater structural and measured hardware savings.}
    \label{fig:OEST}
\end{figure}

Fig.~\ref{fig:OEST} illustrates how structured compression changes clean accuracy, adversarial
robustness, and model complexity non-monotonically. Mathematical compression does not automatically guarantee actual hardware efficiency; thus, relying on a fixed compression ratio is insufficient for dependable edge deployment. Recent BCI studies established that EEG decoders are vulnerable to adversarial manipulation~\cite{zhang2021tiny} and benchmarked multiple defenses~\cite{meng2023benchmark}. 
More recent alignment-based defense benchmarks
demonstrate the benefits of combining preprocessing
and adversarial training for EEG decoding
\cite{chen2025alignment}, but leave
compression-aware operating-point selection
and native hardware costs unexplored.
Prior work, including AERIAL~\cite{rehman2026aerial}, has demonstrated that accuracy-preserving compression does not necessarily preserve adversarial robustness, and that robustness and hardware efficiency may evolve independently. Nonetheless, characterization alone does not establish how to select a physically realizable compressed decoder that satisfies specified accuracy and robustness requirements. Furthermore, the interaction between compression severity and robust recovery remains insufficiently understood, complicating the choice of an efficient deployment configuration. ORBITAL addresses these limitations through robustness-constrained architecture search, recovery-aware evaluation, and native hardware validation.

\begin{figure*}[t]
    \centering
    \includegraphics[width=\linewidth]{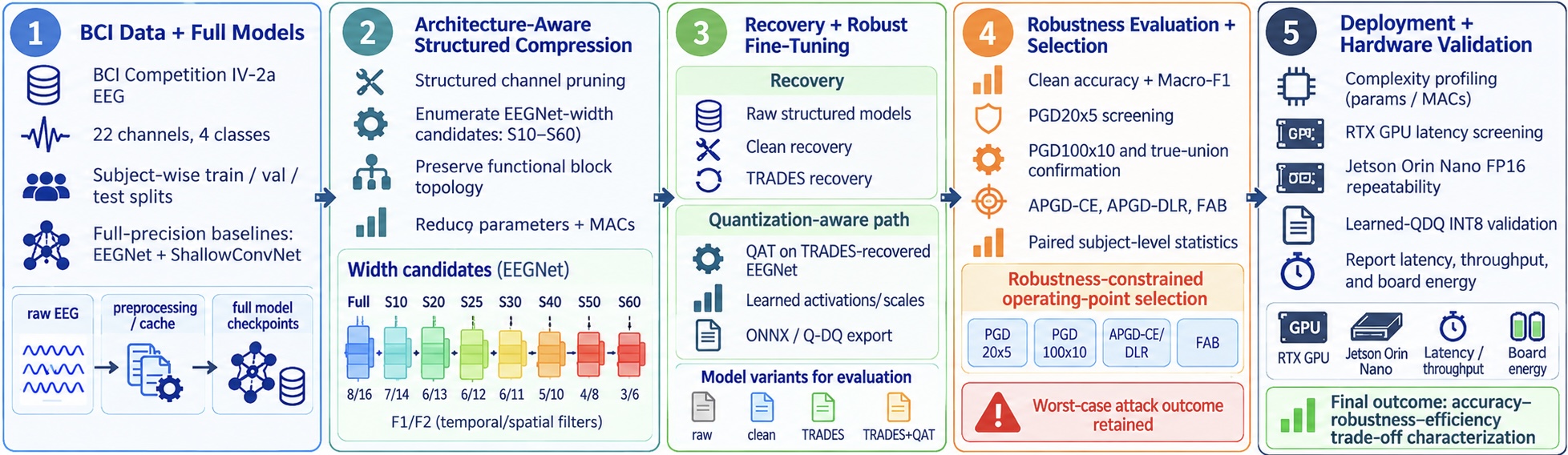}
    \caption{Overview of the ORBITAL methodology. EEGNet serves as the primary deployment architecture, while ShallowConvNet evaluates cross-architecture generalization.}
    \label{fig:meth}
\end{figure*}

Our main contributions are:

\begin{itemize}
    \item \textbf{Robustness-constrained structured
    compression.}
    We introduce a hardware-realizable EEG decoder
    compression workflow that enumerates physically
    valid channel widths, screens clean and adversarial
    performance, and selects constraint-feasible and
    Pareto operating points rather than prescribing
    a fixed compression ratio.

    \item \textbf{Compression--recovery interaction.}
    Through matched full-width TRADES controls and
    compression-level recovery ablations, we show
    that robust training is the principal source of
    adversarial recovery, while its benefit depends
    significantly on compression severity. At S25,
    TRADES improves true-union robustness by
    6.74 percentage points over clean-only recovery
    across all nine subjects.

    \item \textbf{Adversarial boundary characterization.}
    Using per-example robustness transitions,
    post-attack margins, gradient geometry, and
    random-channel controls, we show that compression
    redistributes adversarial robustness rather than
    uniformly increasing classification margins.
    The S25 operating point achieves the highest
    observed mean union robustness among the
    compressed EEGNet candidates.

    \item \textbf{Native edge-hardware validation.}
    We quantify actual latency, board power, and
    energy on Jetson Orin Nano using repeated FP16
    inference measurements. We additionally show
    that learned-scale INT8 Q/DQ can preserve model
    predictions while increasing deployment cost
    because of conversion and kernel overhead.
\end{itemize}

\section{Methodology}
ORBITAL integrates physical structured compression, post-pruning
recovery, robust evaluation, and deployment profiling into a unified
pipeline. Algorithm~\ref{alg:orbital} summarizes the complete procedure. The following subsections detail
each stage.

\subsection{Physical Structured Compression}

ORBITAL performs structured channel pruning to obtain
smaller, dense EEG decoders rather than sparsely masked
networks. For EEGNet, coupled temporal and pointwise
filter dimensions $(F_1,F_2)$ are reduced while preserving
the depthwise structure and downstream compatibility.
Retained parameters are transferred into the rebuilt
architecture. As for ShallowConvNet, filter importance accounts for
filter magnitude and downstream connectivity, with
corresponding convolutional, normalization, and
classifier dimensions rebuilt consistently.
Unlike unstructured sparsity, the resulting models
directly reduce parameter storage and arithmetic
requirements~\cite{li2017pruning,molchanov2017pruning}.

\begin{algorithm}[t]
\caption{ORBITAL: Robustness-Constrained Operating-Point Selection}
\label{alg:orbital}
\footnotesize
\begin{algorithmic}[1]
\REQUIRE Decoder $f$, widths $\mathcal{W}$,
tolerances $\delta_c,\delta_r$
\ENSURE Selected deployable operating points

\STATE Enumerate physically valid widths $\mathcal{W}$
\FOR{$w \in \mathcal{W}$}
    \STATE Rebuild dense decoder $f_w^{\mathrm{raw}}$
    \STATE Recover $f_w^{\mathrm{clean}}$ and $f_w^{\mathrm{TRADES}}$
    \STATE Screen clean and PGD20 accuracy
\ENDFOR
\STATE Average seeds within each subject
\STATE Enforce clean/robust feasibility constraints
\STATE Construct the accuracy--robustness--size Pareto set
\STATE Select feasible and representative candidates
\STATE Confirm with PGD100$\times$10, APGD-CE,
APGD-DLR, and FAB
\STATE Compute clean-correct true-union accuracy
\STATE Profile selected models on edge hardware
\RETURN Constraint-selected, highest-mean-robustness,
and efficiency-oriented points
\end{algorithmic}
\end{algorithm}

\subsection{Robustness-Aware Recovery and Quantization}

Following physical compression, ORBITAL evaluates raw,
clean-recovered, and adversarially recovered decoders.
Robust recovery follows TRADES~\cite{zhang2019trades}:
\begin{equation}
\mathcal{L}_{\mathrm{TRADES}} =
\mathrm{CE}(f(x),y)
+\beta D_{\mathrm{KL}}
\left(p(x)\,\|\,p(x')\right),
\label{eq:trades}
\end{equation}
where $x'$ is an adversarially perturbed EEG input.
An optional parameter-anchor penalty regularizes
deviations from the pruned initialization. For low-precision deployment, EEGNet additionally
supports QAT with fake-quantized activations and
per-output-channel weights. Explicit Q/DQ export
enables evaluation of native TensorRT execution,
separating quantization fidelity from realized
hardware efficiency.

\subsection{Robustness-Constrained Operating-Point Selection}

Rather than selecting compression severity heuristically,
ORBITAL identifies deployable operating points under
explicit clean-accuracy and robustness constraints.
For candidate width $w$, let $A_c(w)$ denote clean
accuracy and $A_r(w)$ denote clean-correct PGD
screening robustness. Relative to the full model
$w_0$, feasibility requires
\begin{equation}
\begin{aligned}
A_c(w)&\geq A_c(w_0)-\delta_c,\\
A_r(w)&\geq A_r(w_0)-\delta_r.
\end{aligned}
\label{eq:feasibility}
\end{equation}

The strict operating point maximizes realized
parameter reduction $R_p(w)$:
\begin{equation}
w^*=\underset{w\in\mathcal{W}_{\mathrm{feas}}}
{\arg\max}\; R_p(w).
\label{eq:selection}
\end{equation}
Nondominated candidates are additionally retained
over clean accuracy, screening robustness, and
parameter reduction. The final robustness is quantified by the intersection
of clean correctness and all confirmation attacks:
\begin{equation}
A_{\mathrm{union}}=
\frac{1}{N}\sum_{i=1}^{N}
\mathbf{1}\left[
f(x_i)=y_i
\land
\bigwedge_{a\in\mathcal{A}}
f(x_i^a)=y_i
\right].
\label{eq:union}
\end{equation}
This prevents initially misclassified trials from
being counted as adversarially robust.

\begin{table*}[t]
\centering
\caption{Strong-attack confirmation across selected
EEGNet operating points on BCI IV-2a.
Results average three seeds within each of nine
subjects. All accuracy values are percentages.
Union requires clean correctness and survival
under all four attacks.}
\label{tab:compression_frontier}
\scriptsize
\setlength{\tabcolsep}{4.0pt}
\begin{tabular}{lccccccccc}
\toprule
\textbf{Variant}
& \textbf{$F_1/F_2$}
& \textbf{Param. $\downarrow$}
& \textbf{MAC $\downarrow$}
& \textbf{Clean}
& \textbf{PGD100}
& \textbf{APGD-CE}
& \textbf{APGD-DLR}
& \textbf{FAB}
& \textbf{Union} \\
\midrule
Full
& 8/16 & -- & -- & \textbf{57.39}
& 22.78 & 22.67 & 21.66 & 21.48 & 21.27 \\

S10-TRADES
& 7/14 & 13.24 & 12.56 & 55.32
& 23.98 & 23.87 & 22.39 & 22.45 & 22.22 \\

\textbf{S25-TRADES}
& \textbf{6/12} & \textbf{26.27} & \textbf{25.10}
& 53.95
& \textbf{24.92} & \textbf{24.76}
& \textbf{23.33} & \textbf{23.15}
& \textbf{23.06} \\

S40-TRADES
& 5/10 & 39.08 & 37.63 & 52.34
& 22.87 & 22.65 & 20.99 & 20.81 & 20.69 \\

S50-TRADES
& 4/8 & 51.68 & 50.14 & 49.83
& 22.58 & 22.29 & 20.19 & 19.87 & 19.70 \\

S60-TRADES
& 3/6 & 64.05 & 62.63 & 44.68
& 20.91 & 20.67 & 18.76 & 18.47 & 18.38 \\
\bottomrule
\end{tabular}
\end{table*}

\section{Experimental Setup}

\textit{\textbf{Dataset and models:}}
Experiments use BCI Competition IV-2a~\cite{tangermann2012bci},
containing nine subjects, 22 EEG channels, and four
motor-imagery classes. Signals are sampled at 250~Hz
and band-pass filtered to 4--38~Hz. Each subject
provides 230/58/288 training/validation/test trials
with 4.5-s input windows. EEGNet is the primary
compression and deployment architecture;
ShallowConvNet evaluates cross-architecture
generalization.

\textit{\textbf{Compression and recovery:}}
We evaluate EEGNet width-reduction targets
$\{0,10,20,25,30,40,50,60\}\%$ with valid
coupled $(F_1,F_2)$ dimensions. Requested width
reductions are distinguished from realized parameter
and MAC reductions. Raw pruning, clean recovery,
and TRADES recovery are evaluated. TRADES uses
$\beta=2$, $\epsilon=0.005$, ten inner steps,
step size $0.00125$, and an anchor coefficient
of $1$. Recovery uses a learning rate of
$5\times10^{-5}$, up to 30 epochs, and
validation-based early stopping.

\textit{\textbf{Selection and robustness evaluation:}}
The screening stage uses PGD20 with one random
restart and predefined clean and robustness
tolerances of $\delta_c=0.03$ and $\delta_r=0.02$.
Selected candidates are confirmed using
PGD100$\times$10, APGD-CE, APGD-DLR, and FAB
under a common $\ell_\infty$ threat radius
$\epsilon=0.005$. Robustness is reported using
the clean-correct true union.

\textit{\textbf{Reproducibility and statistics:}}
EEGNet compression-search and ShallowConvNet
experiments cover nine subjects and three seeds
$\{1,48,550\}$. Seed results are averaged within
each subject before paired subject-level Wilcoxon
tests and bootstrap 95\% confidence intervals.
Selected aggressive EEGNet QAT experiments
remain single-seed.

\textit{\textbf{Hardware evaluation:}}
Native deployment uses TensorRT on NVIDIA
Jetson Orin Nano Super~\cite{nvidiajetson,
nvidiatensorrt} under fixed \textbf{MAXN\_SUPER} clocks.
FP16 configurations are measured in five
independent 60-s runs. We report inference latency,
throughput, active whole-board power
($\mathrm{VDD\_IN}$), and board energy per inference.
Learned-scale Q/DQ exports are separately evaluated
to distinguish numerical fidelity from native
INT8 deployment cost.

\section{Results \& Analysis}
\subsection{Compression--Robustness Operating Points}

Table~\ref{tab:compression_frontier} reports strong-attack confirmation
for selected EEGNet widths. Under the predefined clean-accuracy and
PGD20 screening tolerances, S10-TRADES is the maximum-compression
screen-feasible candidate: it removes 13.24\% of parameters and
12.56\% of MACs while obtaining 55.32\% clean accuracy and 22.22\%
true-union robustness. S25-TRADES achieves the highest observed mean
union robustness among the evaluated compressed candidates (23.06\%)
with 26.27\% fewer parameters and 25.10\% fewer MACs. Its numerical
advantage over the standard full-width decoder (21.27\%) is not
statistically significant (+1.79~pp, $p=0.5703$). Higher compression
reduces both clean and union accuracy, exposing distinct
robustness-oriented and efficiency-oriented operating points.

\subsection{Compression--Recovery Interaction}

\begin{figure}[t]
    \centering
    \includegraphics[width=\columnwidth]{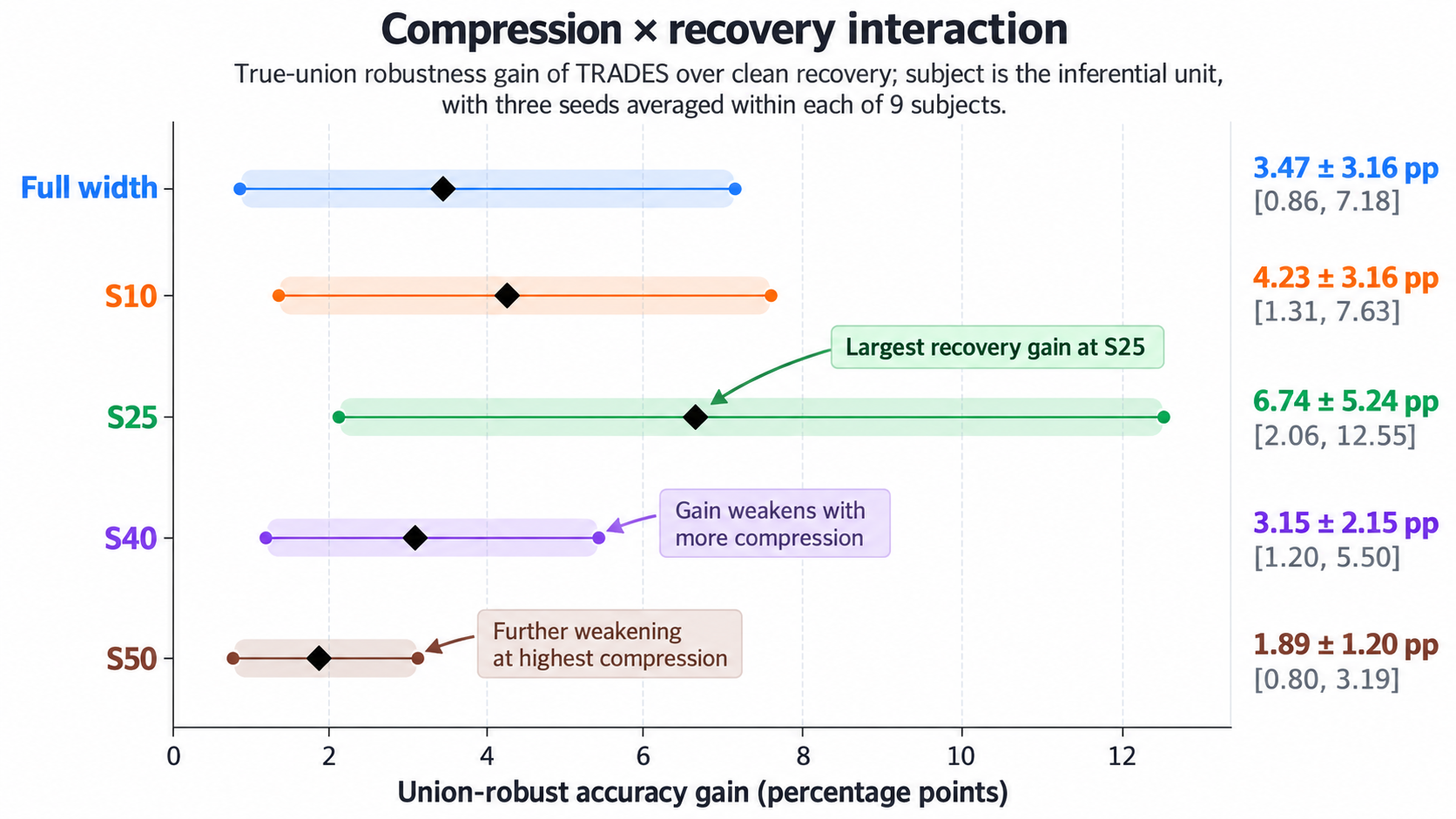}
    \caption{TRADES union-accuracy gain over clean
    recovery across EEGNet compression levels.
    Markers denote subject means; intervals show
    95\% subject-bootstrap confidence intervals.}
    \label{fig:comp}
\end{figure}

Matched full-width controls show that Full-TRADES increases union robustness over standard Full by 3.42~pp ($p=0.0234$), whereas S25-TRADES is numerically 1.63~pp below Full-TRADES ($p=0.0977$). Thus, compression alone does not explain the gain. Fig.~\ref{fig:comp} shows that the TRADES benefit over clean recovery peaks at S25 (+6.74~pp). This recovery gain exceeds that at full width by 3.27~pp ($p=0.0078$), establishing a
compression-dependent recovery interaction. Table~\ref{tab:causal} reports the paired tests.

\begin{figure}[t]
    \centering
    \includegraphics[width=\columnwidth]{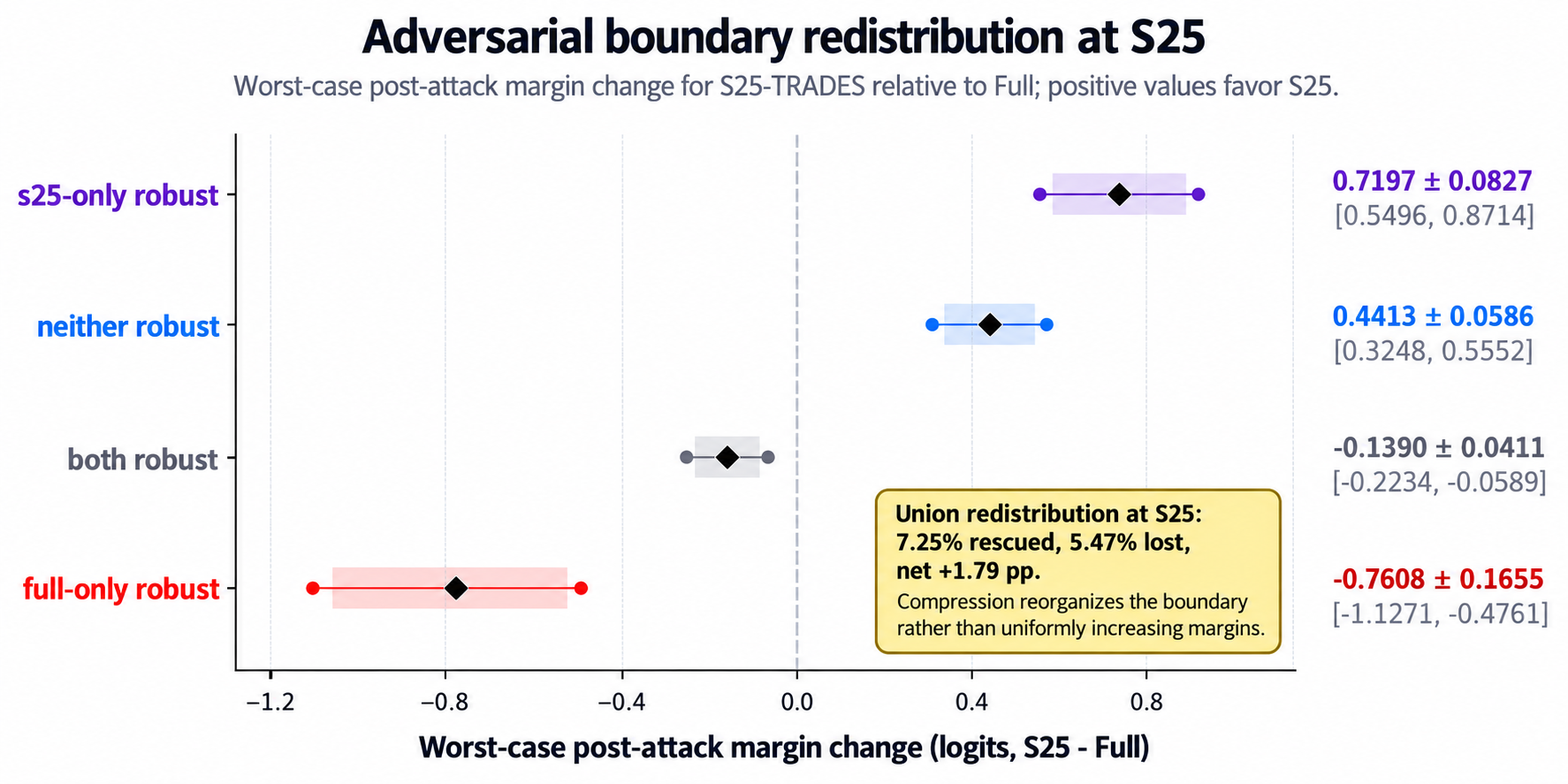}
   \caption{Worst-case post-attack margin differences
between S25-TRADES and standard Full, grouped by
per-example union-robustness outcome.
Intervals denote 95\% confidence intervals.}
    \label{fig:boundary}
\end{figure}

\begin{table}[t]
\centering
\caption{Paired subject-level causal and recovery comparisons.
Union-accuracy differences are in percentage points; $p$ values
are from two-sided Wilcoxon tests after averaging seeds within each
of nine subjects.}
\label{tab:causal}
\scriptsize
\setlength{\tabcolsep}{3.2pt}
\renewcommand{\arraystretch}{1.06}
\begin{tabular}{lrr}
\toprule
\textbf{Comparison} & \textbf{$\Delta$ union} & \textbf{$p$} \\
\midrule
Full-TRADES vs. Full & $+3.42$ & $0.0234$ \\
S25-TRADES vs. Full & $+1.79$ & $0.5703$ \\
S25-TRADES vs. Full-TRADES & $-1.63$ & $0.0977$ \\
S25-TRADES vs. S25-raw & $+7.45$ & $0.0039$ \\
S25-TRADES vs. S25-clean & $+6.74$ & $0.0039$ \\
TRADES gain: S25 vs. Full width & $+3.27$ & $0.0078$ \\
S25 magnitude vs. random & $+0.26$ & $1.0000$ \\
\bottomrule
\end{tabular}
\end{table}

\subsection{Adversarial Boundary Redistribution}

Fig.~\ref{fig:boundary} shows that S25-TRADES
rescues 7.25\% of examples while losing robustness
on 5.47\%, yielding a net +1.79~pp change.
Opposing worst-case margin shifts across rescued
and lost examples indicate selective adversarial
boundary redistribution rather than uniform
margin expansion. Gradient RMS continues declining
beyond S25 despite decreasing union robustness,
so smaller gradients alone do not explain
the observed robustness peak.

\subsection{Cross-Architecture Generalization}

Across nine subjects and three seeds, ShallowConvNet S25-TRADES
achieves 57.56\% clean and 41.23\% union accuracy, compared with
57.90\% and 41.86\% for Full; neither paired difference is significant.
At S50, raw pruning yields 34.84\% union accuracy; clean recovery
restores it to 40.74\%, compared with 39.76\% for S50-TRADES.
Hence, the preferred recovery method depends on architecture and
compression severity rather than following a universal TRADES advantage.

\subsection{Hardware Efficiency and Precision}

\begin{figure}[t]
    \centering
    \includegraphics[width=\columnwidth]{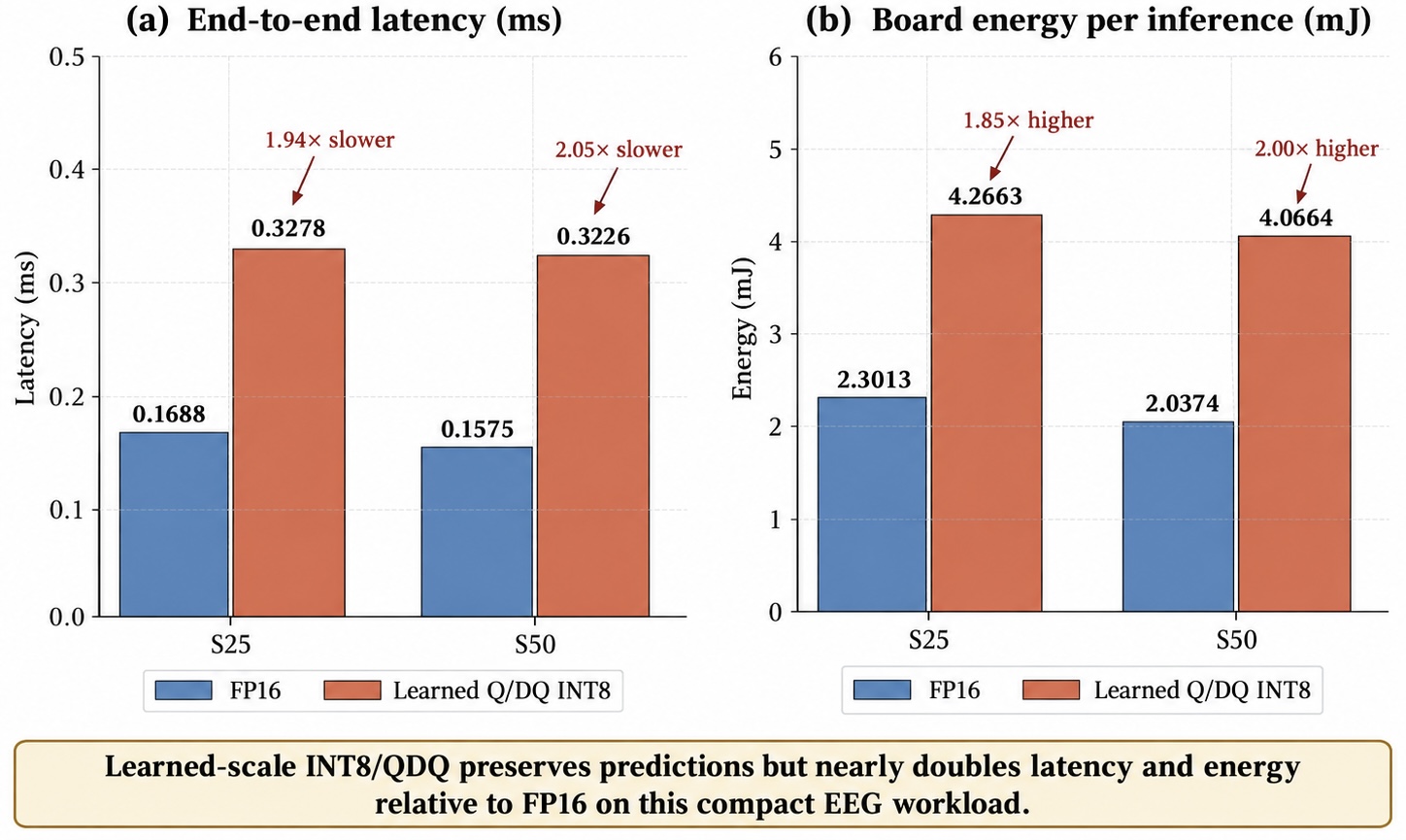}
    \caption{Native Jetson precision comparison. Learned-scale INT8
    Q/DQ preserves approximately 99.3\% prediction agreement with
    the QAT models, yet increases latency and board energy by roughly
    twofold relative to FP16 for S25 and S50.}
    \label{fig:int8_jetson}
\end{figure}

Table~\ref{tab:hw} shows that S25 reduces Jetson board energy by 5.35\% with little latency change, whereas S50 reduces latency and energy by 8.00\% and 16.20\%, respectively. In contrast, Fig.~\ref{fig:int8_jetson}
shows that learned-scale INT8 Q/DQ preserves 99.3\% prediction agreement but incurs roughly twice the FP16 latency and energy. Conversion and reformat overheads therefore limit the benefit of lower precision.

\begin{table}[t]
\centering
\caption{Jetson Orin Nano TensorRT FP16 measurements
(mean $\pm$ SD over five runs).}
\label{tab:hw}
\scriptsize
\setlength{\tabcolsep}{3.2pt}
\begin{tabular}{lrrr}
\toprule
Variant & Latency (ms) & Active power (W) & Energy (mJ) \\
\midrule
Full & $0.1712\!\pm\!0.0029$ & $14.203\!\pm\!0.098$ & $2.4314\!\pm\!0.0244$ \\
S25 & $0.1688\!\pm\!0.0002$ & $13.630\!\pm\!0.017$ & $2.3013\!\pm\!0.0015$ \\
S50 & $0.1575\!\pm\!0.0044$ & $12.940\!\pm\!0.210$ & $2.0374\!\pm\!0.0237$ \\
\bottomrule
\end{tabular}
\end{table}

\section{Conclusion}

To conclude, ORBITAL demonstrates that robust EEG compression
requires joint consideration of structural capacity,
recovery objectives, and deployment behavior.
A constrained search identifies distinct feasible,
robustness-oriented, and efficiency-oriented
operating points. Matched controls attribute the
primary robustness gains to TRADES, while
per-example analysis reveals heterogeneous
adversarial-boundary redistribution.
Cross-architecture experiments confirm that recovery
effectiveness depends on decoder structure and
compression severity. Native Jetson measurements
further show that structural compression can reduce
energy, whereas INT8 Q/DQ may increase execution
cost due to conversion overhead. 

\section*{Acknowledgment}
 This work was supported in part by the NYUAD Center for CyberSecurity (CCS), funded by Tamkeen under the NYUAD Research Institute grant G1104. 
 
\section*{Generative AI Use Disclosure}
The authors used Generative AI tools (specifically ChatGPT and Grammarly) for language editing, text refinement, and visual refinement of the methodology figure. The authors reviewed and edited all generated or refined content as needed and take full responsibility for the publication’s content.
 
\balance
\clearpage
\bibliographystyle{IEEEtran}
\bibliography{references}
\end{document}